\documentclass[letterpaper, 10 pt, conference]{ieeeconf}  

\IEEEoverridecommandlockouts                              

\usepackage{graphicx}
\usepackage{amsmath}
\usepackage{algorithm}
\usepackage{amssymb}
\usepackage{booktabs}

\usepackage{bigstrut}
\usepackage{bbm}
\usepackage{bbding}
\usepackage{bbold}
\usepackage{mathtools}

\usepackage{caption}
\usepackage{subcaption}
\usepackage{multirow}
\usepackage{xcolor}
\usepackage{tcolorbox}
\usepackage{mathptmx}
\usepackage{wrapfig}
\usepackage{hyperref}

\newcommand{\lmp}{\textsc{LLM+P}}
\newcommand{\lmsp}{\textsc{LLM-as-P}}
\newcommand{\lmtot}{\textsc{LLM-as-P (ToT)}}

\newcommand{\blocksworld}{\textsc{Blocksworld}}
\newcommand{\barman}{\textsc{Barman}}
\newcommand{\floortile}{\textsc{Floortile}}
\newcommand{\grippers}{\textsc{Grippers}}
\newcommand{\storage}{\textsc{Storage}}
\newcommand{\termes}{\textsc{Termes}}
\newcommand{\tyreworld}{\textsc{Tyreworld}}

\definecolor{commentcolor}{RGB}{110,154,155}
\definecolor{inputcolor}{RGB}{102, 59, 12}
\definecolor{problemcolor}{RGB}{242,242,242}
\definecolor{resultcolor}{RGB}{252,228,215}
\definecolor{contextcolor}{RGB}{31,160,171}

\newcommand{\PlanCode}[1]{\ttfamily\textcolor{inputcolor}{ #1}}  

\newcommand{\Context}[1]{\colorbox{contextcolor}{\parbox{\linewidth}{#1}}
}

\definecolor{MyDarkRed}{rgb}{0.70980392, 0.09019608,0.0}
\definecolor{MyLightBlue}{rgb}{0.34117647, 0.75686275, 1.}
\definecolor{MyLightGreen}{rgb}{0.5372549 , 0.98431373, 0.30588235}

\title{\LARGE \bf
\lmp{}: Empowering Large Language Models \\with Optimal Planning Proficiency}

\author{
Bo Liu$^{*\dagger}$, Yuqian Jiang$^{*\dagger}$, Xiaohan Zhang$^{\ddagger}$, Qiang Liu$^{\dagger}$, Shiqi Zhang$^{\ddagger}$, Joydeep Biswas$^{\dagger}$, Peter Stone$^{\dagger\mathsection}$
\thanks{*Equal contribution.}
\thanks{$^\dagger$Department of Computer Science, The University of Texas at Austin \texttt{\{bliu, lqiang, joydeep, pstone\}@cs.utexas.edu, jiangyuqian@utexas.edu}}
\thanks{$^\ddagger$Department of Computer Science, State University of New York at Binghamton \texttt{\{xzhan244,zhangs\}@binghamton.edu }}
\thanks{$^\mathsection$Sony AI}
}

\begin{document}
\maketitle
\begin{abstract}
Large language models (LLMs) have demonstrated remarkable zero-shot generalization abilities: state-of-the-art chatbots can provide plausible answers to many common questions that arise in daily life.
However, so far, LLMs cannot reliably solve long-horizon robot planning problems.
By contrast, classical planners, once a problem is given in a formatted way, can use efficient search algorithms to quickly identify correct, or even optimal, plans.
In an effort to get the best of both worlds, this paper introduces \lmp{}, the first framework that incorporates the strengths of classical planners into LLMs. \lmp{} takes in a natural language description of a planning problem, then returns a correct (or optimal) plan for solving that problem in natural language. \lmp{} does so by first converting the language description into a  file written in the planning domain definition language (PDDL), then leveraging classical planners to quickly find a solution, and then translating the found solution back into natural language. Along with \lmp{}, we define a diverse set of different benchmark problems taken from robot planning scenarios. 
Via a comprehensive set of experiments on these benchmark problems, we find that \lmp{} is able to provide \emph{optimal} solutions for most problems, while LLMs fail to provide even feasible plans for most problems. 
We also show \lmp{} enables a home robot to solve a complex manipulation task that is specified by the user in natural language.
\footnote{The code and results are publicly available at \url{https://github.com/Cranial-XIX/llm-pddl.git}.}
\end{abstract}



\section{Introduction}
\label{sec::intro}
Ever since the birth of the field, AI researchers have sought to create programs that can converse in natural language with the same grace and flexibility as people.   While even relatively simple models, such as Eliza from 1966~\cite{weizenbaum1966eliza}, can generate responses to some prompts that seem reasonable, it has always been relatively easy to generate prompts that expose their weaknesses compared to people --- their lack of true ``understanding." 
 
While large language models (LLMs) such as GPT-4~\cite{openai2023gpt4} and ChatGPT~\cite{vemprala2023chatgpt} have far surpassed expectations of just a few years ago, they are no different in this respect.  Indeed the internet is now awash with examples of people reveling in getting ChatGPT to generate output that even a 5-year-old human child would know to be ill-advised.

\begin{figure*}
    \centering
    \includegraphics[width=0.8\textwidth]{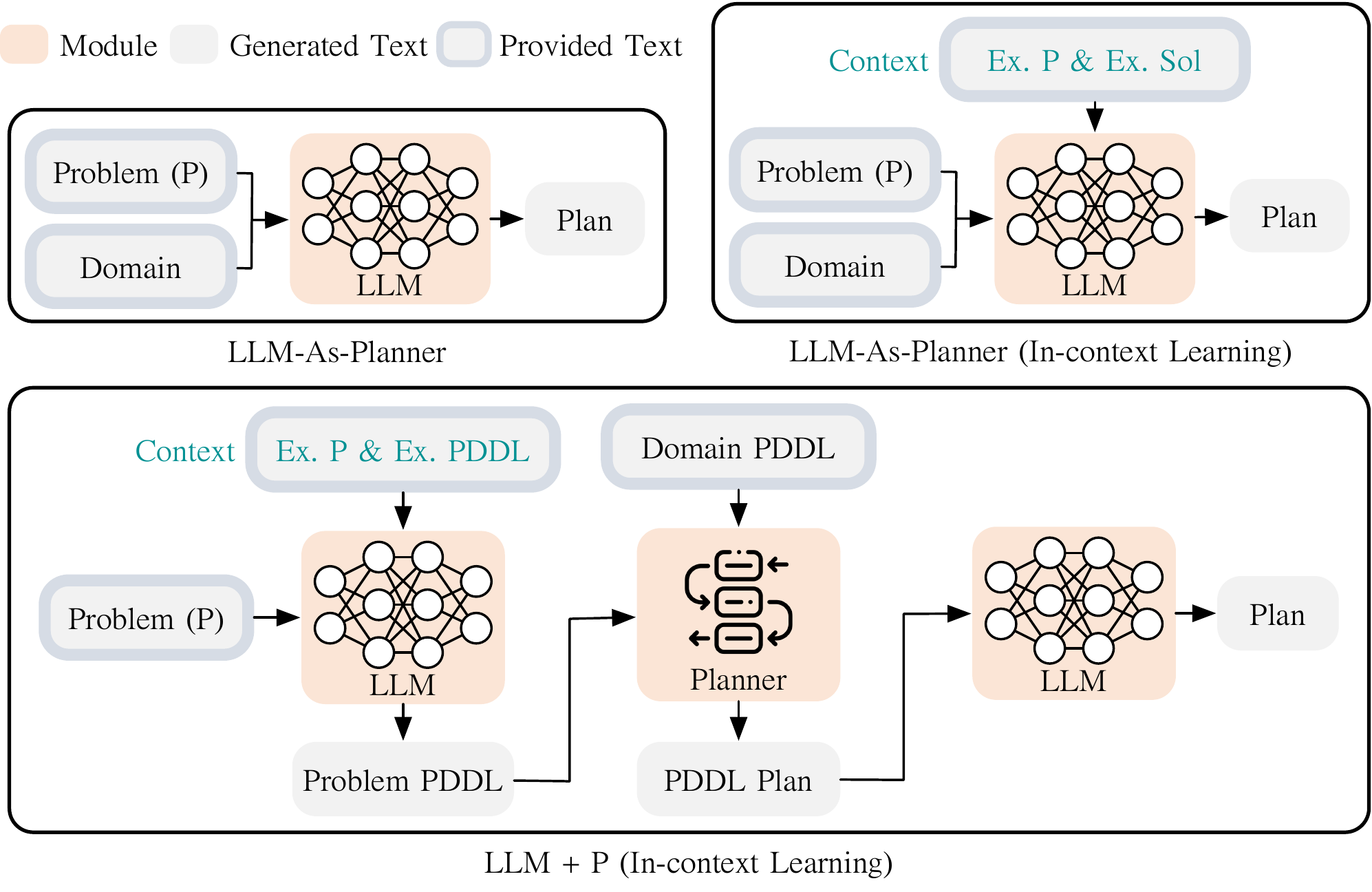}
    \caption{\lmp{} makes use of a large language model (LLM) to produce the PDDL description of the given problem, then leverages a classical planner for finding an \emph{optimal} plan, then translates the raw plan back to natural language using the LLM again.}
    \label{fig:intro}
\end{figure*}

Given how LLMs are designed and trained, this phenomenon should come as no surprise.  
 They are not specifically built to demonstrate understanding.  They are trained to generate sequences of words that \emph{might be plausible to a human} given a prior context. 
In the terms of Mahowald et al.~\cite{mahowald2023dissociating} LLMs have become amazingly proficient at \emph{linguistic competence} --- knowing \emph{how} to say things;
   but they are not nearly as good at \emph{functional competence} ---
   knowing \emph{what} to say.  Specifically, they can be (relatively) easily fooled by, for example, asking for the result of a straightforward arithmetic problem that does not appear in their training corpus 
or by asking for the solution to a novel planning problem that requires knowledge about how the world works. 

\begin{tcolorbox}[
    standard jigsaw,
    title=A Failure Example of GPT-4 in Planning,
    opacityback=0,
]
\textbf{Problem (P1):} You have 5 blocks. One cannot place more than one block on another block.
b5 is on top of b3. 
b4 is on top of b2. 
b2 is on top of b1. 
b3 is on top of b4. 
b1 is on the table. 
b5 is clear. 
Your arm is empty. 

Your goal is to move the blocks. 

b1 should be on top of b2. 

b3 should be on top of b5. 

b4 should be on top of b1. 
\\\\
\textbf{GPT-4:}
\\
\PlanCode{
1. Move b5 from b3 to the table. \\
2. Move b3 from b4 to b5. \\
3. Move b4 from b2 to the table. \\
4. Move b2 from b1 to b5. \\
5. Move b1 from the table to b2. \\
6. Move b4 from the table to b1.}
\begin{wrapfigure}{r}{0.35\textwidth}
  \centering
  \vspace{-160pt}
  \includegraphics[width=0.35\textwidth]{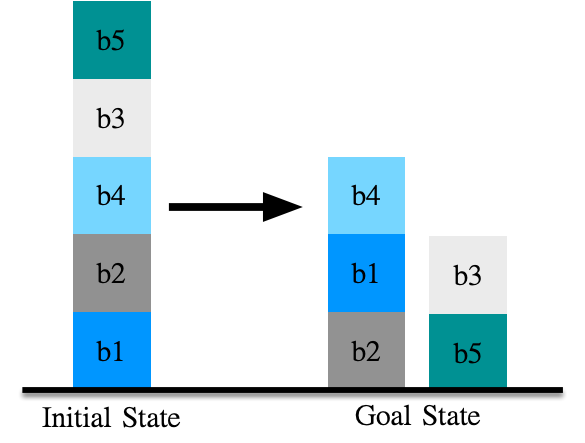}
\end{wrapfigure}
\end{tcolorbox}
 
Does that mean that we should increase efforts to include all arithmetic and planning problems in their training corpus?  Clearly, that is a fool's errand. On the other hand, why should it be necessary?  We already have calculators and general-purpose symbolic planners that are guaranteed to produce correct answers.  Thus a natural alternative approach, and one that we are admittedly not the first to explore, is to connect LLMs to such tools. 
 
With this motivation in mind, the objective of the research reported in this paper is, for the first time,  to enable LLMs to solve planning problems \emph{correctly}.  We aim to do so without altering the LLMs themselves, even with finetuning~\cite{lee2019mixout, wei2021finetuned}.  Rather, we introduce a methodology, called \lmp{} by which, when posed a natural language description of a planning problem, the LLM: 
\begin{enumerate} 
  \item outputs a problem description suitable as input to a general-purpose planner; 
  \item solves the problem using the general-purpose planner; and  
  \item converts the output of the planner back to natural language (or connects to action executors of a robot).
\end{enumerate} 
 
Our extensive empirical evaluations indicate that \lmp{} is able to generate correct solutions to many more planning problems than are LLMs on their own. While demonstrated in this paper on planning problems, this general methodology can be applied to any class of problems for which we have a sound and complete solver, such as arithmetic problems (by leveraging calculators). 

\noindent\textbf{Limitation:} In this paper, we do not
ask the LLM to \emph{recognize} that it has been posed a prompt that is
suitable for processing using the proposed \lmp{} pipeline. A valuable future research direction will be to consider recognizing when a prompt should be processed by LLM+P.
 

\section{Background}
\label{sec::background}

This section introduces the notation we use for representing a planning problem to be solved by LLMs, and recaps the standard representation of classical planners.

\subsection{The Classical Planning Problem}
Formally, the input of a planning problem $P$ is defined by a tuple $\langle \mathcal{S}, s^{init}, \mathcal{S}^{G}, \mathcal{A}, f\rangle$:
\begin{itemize}
    \item $\mathcal{S}$ is a finite and discrete set of states used to describe the world's state (i.e., state space).
    We assume a factored state space such that each state $s \in \mathcal{S}$ is defined by the values of a fixed set of variables.
    \item $s^{init} \in \mathcal{S}$ is an initial world state.
    \item $\mathcal{S}^{G} \subset \mathcal{S}$ is a set of goal states. $\mathcal{S}^{G}$ are usually specified as a list of \emph{goal conditions}, all of which must hold in a goal state.
    \item $\mathcal{A}$ is a set of symbolic actions. 
    \item $f$ is the underlying state transition function. $f$ takes the current state and an action as input and outputs the corresponding next state. 
    
\end{itemize}
A solution to a planning problem $P$ is a symbolic plan $\pi$ in the form of $\langle a_1, a_2, \dots, a_N \rangle$, such that the preconditions of $a_1$ hold in $s^{init}$, the preconditions of $a_2$ hold in the state that results from applying $a_1$, and so on, with the goal conditions all holding in the state that results after applying $a_N$.

\subsection{Planning Domain Definition Language (PDDL)}
The planning domain definition language (PDDL) serves as a standardized encoding of classical planning problems~\cite{mcdermott1998pddl, haslum2019introduction}. 
The PDDL representation of a planning problem $P$ is separated into two files: a domain file and a problem file. The domain PDDL file provides a lifted representation of the underlying rules of the world. It includes a set of predicates that define the state space $\mathcal{S}$ and the actions (i.e., $\mathcal{A}$) with their preconditions and effects (i.e., the transition function $f$). The problem PDDL file provides a list of objects to ground the domain, the problem's initial state $s^{init}$ and goal conditions $\mathcal{S}^{G}$. 
There exists a rich set of symbolic planners that implement efficient search algorithms to solve planning problems formalized in PDDL.
In this work, we aim to take a natural language prompt which describes the initial state $s^{init}$ and goal conditions $\mathcal{S}^{G}$, formulate it in PDDL, and leverage symbolic planners to output correct plans. We assume the domain rules are available (See the assumptions in Section~\ref{sec::method}).

\vspace{-15pt}
\section{Method}
\label{sec::method}
\vspace{-5pt}
This section first uses a running example to investigate ways for formulating planning prompts in PDDL and then introduces the \lmp{} method.
\vspace{-10pt}
\subsection{Large Language Model as a PDDL Writer}
LLMs are bad at planning (or long-horizon reasoning) \cite{valmeekam2022large} but they are good at describing and translating textual inputs, including re-writing planning prompts in the PDDL format. 
The intuition is that we can view PDDL as a different language than English, so re-writing planning prompts in PDDL is essentially a ``machine translation" task that LLMs are known to excel at.
Here we provide an example of a PDDL problem file written by GPT-4 without any prompt-engineering. As we see, the generated file appears to have the correct PDDL syntax but uses a made-up predicate (empty) and misses the initial condition that \texttt{b1} is on the table.

\begin{tcolorbox}[
    standard jigsaw,
    title=An Example PDDL Problem File Written by GPT-4,
    opacityback=0,
]
\textbf{Prompt:} \textcolor{purple}{Description of Problem (P1)} +
Provide me with the problem PDDL file that describes the planning problem directly without further explanations. \\\\
\textbf{GPT-4 (the generated problem PDDL):}

\PlanCode{(:objects b1 b2 b3 b4 b5 - block)} \\
\PlanCode{(:init (on b5 b3) (on b4 b2) (on b2 b1) (on b3 b4) (clear b5) (empty)) }\\
\PlanCode{(:goal (and (on b1 b2) (on b3 b5) (on b4 b1)))}
\end{tcolorbox}
\vspace{-10pt}
\subsection{In-Context Learning}
\vspace{-5pt}
LLMs are known to be capable of in-context learning without finetuning their parameters. By in-context learning, we mean LLMs' ability to perform unseen downstream tasks by simply conditioning on a few input-label pairs (demonstrations)~\cite{brown2020language}.
Here is a pair of problem descriptions in natural language and in PDDL provided to GPT-4 as context. When the context is included with the prompt from the example above, the resulting PDDL problem file is directly solvable by the planner.

\begin{tcolorbox}[
    standard jigsaw,
    title=An Example PDDL Problem File Written by GPT-4 with Context,
    opacityback=0,
]
\textbf{Context (shaded):}

\Context{An example planning problem is: 
 You have 5 blocks. 
b2 is on top of b5. 
b5 is on top of b1. 
b1 is on top of b4. 
b3 is on top of b2. 
b4 is on the table. 
b3 is clear. 
Your arm is empty. 
Your goal is to move the blocks. 
b4 should be on top of b3. The problem PDDL is: 

{\PlanCode{(:objects b1 b2 b3 b4 b5)}\\
\PlanCode{(:init (arm-empty) (on b1 b4) (on b2 b5) (on b3 b2) }\\
\PlanCode{\qquad~~ (on-table b4) (on b5 b1) (clear b3))} \\
\PlanCode{(:goal (and (on b4 b3))))}}}

\textbf{Prompt:} \textcolor{purple}{Description of Problem (P1)} + \textcolor{purple}{Context} +
Provide me with the problem PDDL file that describes the planning problem directly without further explanations. \\


\textbf{GPT-4 (the generated problem PDDL):}

{\PlanCode{
(:objects b1 b2 b3 b4 b5 )}\\
\PlanCode{(:init (arm-empty) (on-table b1) (on b2 b1) (on b3 b4)  }\\
\PlanCode{\qquad ~~~(on b4 b2) (on b5 b3) (clear b5)))} \\
\PlanCode{(:goal (and (on b1 b2) (on b3 b5) (on b4 b1)))}\\\\
}
\textbf{Planner:}\\
{\PlanCode{(unstack b5 b3)
(putdown b5)
(unstack b3 b4)
(stack b3 b5)
}\\
\PlanCode{(unstack b4 b2) (putdown b4) (unstack b2 b1) (putdown b2)}\\
\PlanCode{(pickup b1) (stack b1 b2) (pickup b4) (stack b4 b1)}}
\end{tcolorbox}

\subsection{Large Language Model + Classical Planner (\lmp{})}

Having introduced the LLM's ability to encode problems in PDDL and in-context learning, we are ready to introduce the proposed \lmp{} solution (the bottom of Fig.~\ref{fig:intro}). In particular, we assume that for each problem domain, a human expert can provide a domain description (i.e. action preconditions and effects) that will be fixed for all problem instances that happen in that domain. While the problem of automatically generating the description is another valuable research question, in this proposed work, we assume that the description is available as a PDDL domain file. 
The \lmp{} method is directly applicable as a natural language interface for giving tasks to robot systems. 
For instance, assume we want a robot to act as a bartender to make cocktails. It is reasonable to tell it what actions it can take, but leave itself to infer how to make new cocktails most efficiently given a set of ingredients to combine. 
Moreover, we assume the agent is provided with a \emph{minimal} example that demonstrates what an example problem PDDL looks like for a simple problem inside that domain. Next, the agent is provided with a new (potentially quite complicated) problem ($P$). The LLM then uses the in-context learning to infer the problem PDDL file corresponding to $P$. Once the problem PDDL file is generated, we feed it into any classical planner, together with the provided domain PDDL file, to generate a PDDL plan~\cite{helmert2006fast}. 
In the end, the LLM translates the PDDL plan back into the natural language to finish up the \lmp{} pipeline. \\
\textbf{\textcolor{purple}{To summarize, the assumptions we need for \lmp{} are:}}
\begin{enumerate}
    \item A robot knows when to trigger \lmp{} based on its conversation with a human user.
    \item A domain PDDL is provided to define the actions that the robot is capable of. This specification is task-agnostic --- the entities relevant to the task are specified in the LLM-generated problem PDDL.
    \item A simple problem description in natural language and its corresponding problem PDDL file are also provided.
\end{enumerate}
\section{Related Work}
\label{sec::related}
This section first provides a brief overview of classical planning algorithms.  Then it summarizes recent advances in using large language models for planning tasks.
It concludes with a discussion of recent research on augmenting LLMs with external modules.

\subsection{Classical Planning} 
Automated planning~(or classical planning) techniques can be used for computing a sequence of actions that achieves a given goal~\cite{bylander1994ComputationalComplexityPropositional,mccarthy1963situations,fikes1971strips}.
Automated planning algorithms have been widely used in robot systems. 
Shakey is the first robot that was equipped with a planning component, which was constructed using STRIPS~\cite{nilsson1984shakey}. 
Some previous general-purpose planning architectures were also demonstrated to be useful for robot planning, such as PRODIGY~\cite{carbonell1991prodigy} and HTN~\cite{nau2003shop2}.
Recent classical planning systems designed for robotics frequently use planning domain description language~(PDDL) or answer set programming~(ASP) as the underlying action language for the planners~\cite{jiang2019task, brewka2011answer, lifschitz2002answer, fox2003pddl2}.
For example, researchers have used classical planning algorithms for sequencing actions for a mobile robot working on delivery tasks~\cite{zhang2015mobile}, reasoning about safe and efficient urban driving behaviors for autonomous vehicles~\cite{ding2020task}, and planning actions for a team of mobile robots~\cite{jiang2019multi}. 
Task and motion planning~(TAMP) is a hierarchical planning framework that combines classical planning in discrete spaces and robot motion planning in continuous space~\cite{lagriffoul2018platform, kaelbling2013integrated}.

Most of the above-mentioned planning methods require domain-specific programming languages as the underlying representation of the problems and their solutions.
\lmp{}, on the other hand, takes advantage of LLMs and serves as a natural language interface for robots to solve complex planning tasks. 
The main feature that motivates us to use such classical planning systems is that most of these planners are sound and complete, meaning that they are guaranteed to be logically correct and will output a plan if one exists.  Many are also able to find optimal (shortest) plans, at least if given sufficient time.

\subsection{Planning with Large Language Models}
Various large language models~(LLMs) have been developed in recent years, such as Bert~\cite{devlin2018bert}, CodeX~\cite{chen2021evaluating}, Opt~\cite{zhang2022opt}, GPT-3~\cite{brown2020language}, ChatGPT~\cite{openai}, GPT-4~\cite{openai2023gpt4}, Llama~\cite{touvron2023llama}, Llama2~\cite{touvron2023llama2}, and PaLM~\cite{chowdhery2022palm}.
As LLMs are pretrained with a tremendous amount of offline text data, they can emerge with surprising zero-shot generalization ability, which can be leveraged for robot planning tasks~\cite{ahn2022can,ding2023task,driess2023palm,huang2022inner,huang2022language,kant2022housekeep,singh2022progprompt, lin2023text2motion, yang2023automatonbased, ding2023integrating, ren2023robots, chen2023autotamp}. Several recent methods had successes in extracting task knowledge from LLMs to decompose commands or instructions for robots in natural language.
For instance, the work of Huang et al. showed that LLMs can be used for task planning in household domains by iteratively augmenting prompts~\cite{huang2022language}.
SayCan is another approach that enabled robot planning with affordance functions to account for action feasibility, where the service requests are specified in natural language~\cite{ahn2022can}.
Vemprala et al. recently studied how ChatGPT can be applied to generalized robotics domains~\cite{vemprala2023chatgpt}. 

However, a major drawback of existing LLMs is their lack of long-horizon reasoning ability for complex tasks (See \cite{valmeekam2022large, valmeekam2023planning} and Section 8.2 from \cite{openai2023gpt4}). 
Specifically, the output they produce when presented with such a task is often incorrect in the sense that following the output plan will not actually solve the task.  
Therefore, in this work, we focus on resolving this issue by leveraging the properties of classical planners. 
Similarly, some recent work also investigates approaches for combining classical planning with LLMs~\cite{silver2022pddl, pallagani2022plansformer, arora2023learning, guan2023leveraging, silver2023generalized, pallagani2023understanding, valmeekam2023planningnew, xie2023translating, hazra2023saycanpay, rana2023sayplan, zhou2023isr}. 
They either use prompting or fine-tuning to make LLMs capable of solving PDDL planning problems. Improvements to long-horizon planning capabilities have also been made by iteratively querying LLMs, as demonstrated in Minecraft~\cite{wang2023describe}.
In contrast, we do not solely rely on LLM as the problem solver, but are more into taking the advantage of both the planner~(i.e., generating accurate and optimal plans) and the LLM itself~(i.e., 1-shot generalization for translating natural-language problem descriptions into PDDL).

\subsection{Augmenting LLMs with External Modules}
Recently developed methods have shown that the performance of downstream tasks of LLMs can be improved by combining them with external modules.
For instance, WebGPT~\cite{nakano2021webgpt} is a fine-tuned version of GPT-3 by combining web knowledge to answer open-ended questions.
Lazaridou et al. studied how search engines like Google can be utilized as external tools for LLMs~\cite{lazaridou2022internet}.
MemPrompt~\cite{madaan2023memoryassisted} presented a human-in-the-loop system where a growing memory of errors and user feedback is served as past experience adding to the prompts for more accurately answering new questions.
REPLUG~\cite{shi2023replug} is another retrieval-augmented language modeling paradigm that treats the language model as a black box and augments it with a tuneable retrieval model. Specifically, people have investigated using calculators for computation~\cite{chen2022program,gao2022pal}.
In very recent work related to ours, Schick et al. trained a model called ToolFormer that can decide when and how to call certain tool APIs by in-line augmentation on prompts for LLMs~\cite{schick2023toolformer}. 
In this work, we propose that classical planners can be another particularly useful external module.
In comparison, \lmp{}, does not rely on any fine-tuning or re-training of LLMs. By simply incorporating knowledge from classical planners, \lmp{} incorporates long-horizon reasoning and planning capabilities into existing LLMs. 

The authors are informed that a concurrent work~\cite{lyu2023faithful} presents preliminary results of integrating LLMs with PDDL using the SayCan dataset~\cite{ahn2022can}. However, the SayCan dataset has a limited scope, as it contains only three predefined actions. Consequently, all model variants evaluated in the original paper achieved a success rate of approximately 90\%. Due to the homogeneity of the SayCan dataset, Lyu et al. did not necessitate a rigorous definition of the domain PDDL, which can lead to infeasible plans. As a result, we consider our \lmp{} method as a more comprehensive investigation into enhancing LLMs with optimal planning proficiency.


\section{Experiments}
\label{sec::exp}
We conduct experiments to answer these questions:
\begin{enumerate}
    \item How well does \lmsp{} work?  To what extent can state-of-the-art LLMs and LLM-based reasoning methods be directly used for planning? \textbf{\textcolor{purple}{(Not at all)}}
    \item How well does \lmp{} work compare to \lmsp{}? \textbf{\textcolor{purple}{(Much better)}}
    \item What role does the context play in the success of \lmp{}? \textbf{\textcolor{purple}{(It's crucial)}}
    \item Can \lmp{} help make service robots more efficient on realistic tasks? \textbf{\textcolor{purple}{(Yes)}}
\end{enumerate}
\vspace{-5pt}
\subsection{Benchmark Problems}
We present seven robot planning domains borrowed from past International Planning Competitions and 20 automatically generated tasks for each domain \cite{seipp-et-al-zenodo2022}. Below is a list of the planning domains, along with a brief summary of each.
\begin{enumerate}
    \item \blocksworld{}: Given a set of piles of blocks on a table, a robot is tasked with rearranging them into a specified target configuration. 
    \item \barman{}: A robot bartender is tasked with creating cocktails for a customer's order, utilizing the available ingredients and containers.
    \item \floortile{}: 
    A set of robots are tasked to use paint color patterns on floor tiles. Robots can move around and change colors but cannot step on painted tiles.
    \item \grippers{}: A set of robots with two grippers is given a task to move objects among different rooms.
    \item \storage{}: Given a set of hoists, the goal is to lift and drop crates using the hoists into a depot. Crates are initially stored in different areas and hoists can be moved among storage areas.  
    \item \termes{}: A robot is tasked to build complex structures by carrying and placing blocks, and also climbing on them so that it can build towers. 
    \item \tyreworld{}: The robot is given a task to replace flat tires by, for example, inflating tires, tightening nuts, and moving tools back to the boot when done, all in the proper order.
\end{enumerate}
For each problem $P$, $P$ comes with a natural language description and a ground-truth problem PDDL file. Each domain also includes an example problem description, a corresponding PDDL file, and a plan description, used as context in various approaches. We assume each problem domain has its own domain PDDL file given by the user or a domain expert prior to addressing any planning problems in that domain. This dataset is made publicly available in our codebase for reproducibility.

\subsection{Experiment Setup}
We leverage the \textsc{gpt-4} model provided by OpenAI\footnote{We use the most recent model as of September 2023. \url{https://platform.openai.com/docs/models/gpt-4}} 
for all experiments. We set the temperature to $0$, and use the top probability response. As a result, the response returned from the LLM is  deterministic. Once a text PDDL response is generated, we feed it into the \textsc{fast-downward} planner\footnote{\url{https://github.com/aibasel/downward/tree/release-22.12.0}} and try both aliases \textsc{seq-opt-fdss-1} (guaranteed optimal) and \textsc{lama} (not guaranteed optimal) with a maximum search time of $200$ seconds. We report the success rate of the optimal alias, and for the domains that time out, we show the success rate of the sub-optimal alias in parentheses. For the baseline methods, we manually count the number of optimal plans, and report the number of correct plans in parentheses (if there are any sub-optimal plans). 

We also evaluate a recent LLM-based approach for deliberate reasoning called Tree of Thoughts \cite{yao2023tree}, referred to as \lmtot{}. We adapt the breadth-first-search algorithm from the original ToT implementation\footnote{\url{https://github.com/princeton-nlp/tree-of-thought-llm/}} for planning. The LLM is prompted to expand the search tree from allowed actions and evaluate the paths on their likelihood of reaching the goal. The same time limit of $200$ seconds is applied.


\subsection{Results and Analysis}
The results of applying \lmsp{} and \lmp{} across 7 domains are provided in Table~\ref{tab:main-table}.


\begin{table}[h!]
    \centering
    \scriptsize
    \captionsetup{font=small}
    \begin{tabular}{ l  r r r r r}
    \toprule 
    \multirow{2}{*}{Domain} & \multicolumn{5}{c}{Success Rate \%} \\
    \cmidrule(lr){2-6}
                               & LLM$^-$ & LLM & LLM$^{ToT}$ & \lmp{}$^-$ & \lmp{} \\
    \midrule
    \barman{}~~~~~             & 0        & 0                        &  0           & 0              & \textbf{20 (100)}        \\
    \blocksworld{}             & 20       & 15 (30)                        &  0 (5)            & 0              & \textbf{90}        \\
    \floortile{}               & 0        & 0                        &  0            & 0              & \textbf{0}        \\
    \grippers{}                & 25 (60)     & 35 (50)     &  10 (20)  & 0              & \textbf{95 (100)}        \\
    \storage{}                 & 0        & 0  (25)                      &  0            & 0              & \textbf{85}        \\
    \termes{}                  & 0        & 0                        &  0            & 0              & \textbf{20}        \\
    \tyreworld{}               & 5        & 15            &  0            & 0              & \textbf{10 (90)}        \\
    \bottomrule
    \end{tabular}
\caption{Success rate \% of applying \lmsp{} without context (LLM$^-$), \lmsp{} (LLM), Tree of Thoughts (LLM$^{ToT}$), \lmp{} without context (LLM$^-$), and \lmp{}. 
}
\label{tab:main-table}
\vspace{-5pt}
\end{table}

\begin{figure*}[h]
     \centering
     \begin{subfigure}[b]{0.15\textwidth}
         \centering
         \includegraphics[width=\textwidth]{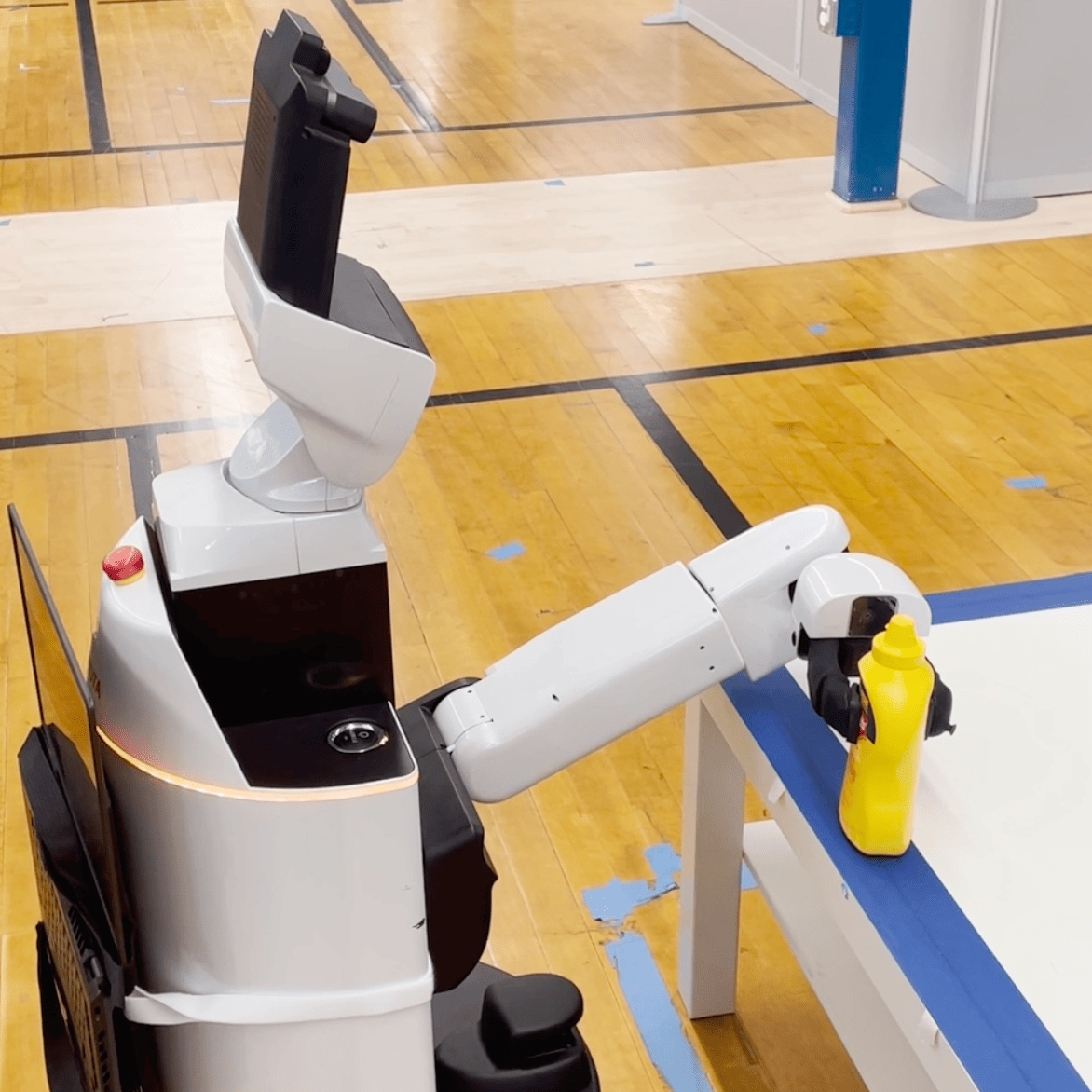}
         \caption{grasp bottle}
     \end{subfigure}
     \begin{subfigure}[b]{0.15\textwidth}
         \centering
         \includegraphics[width=\textwidth]{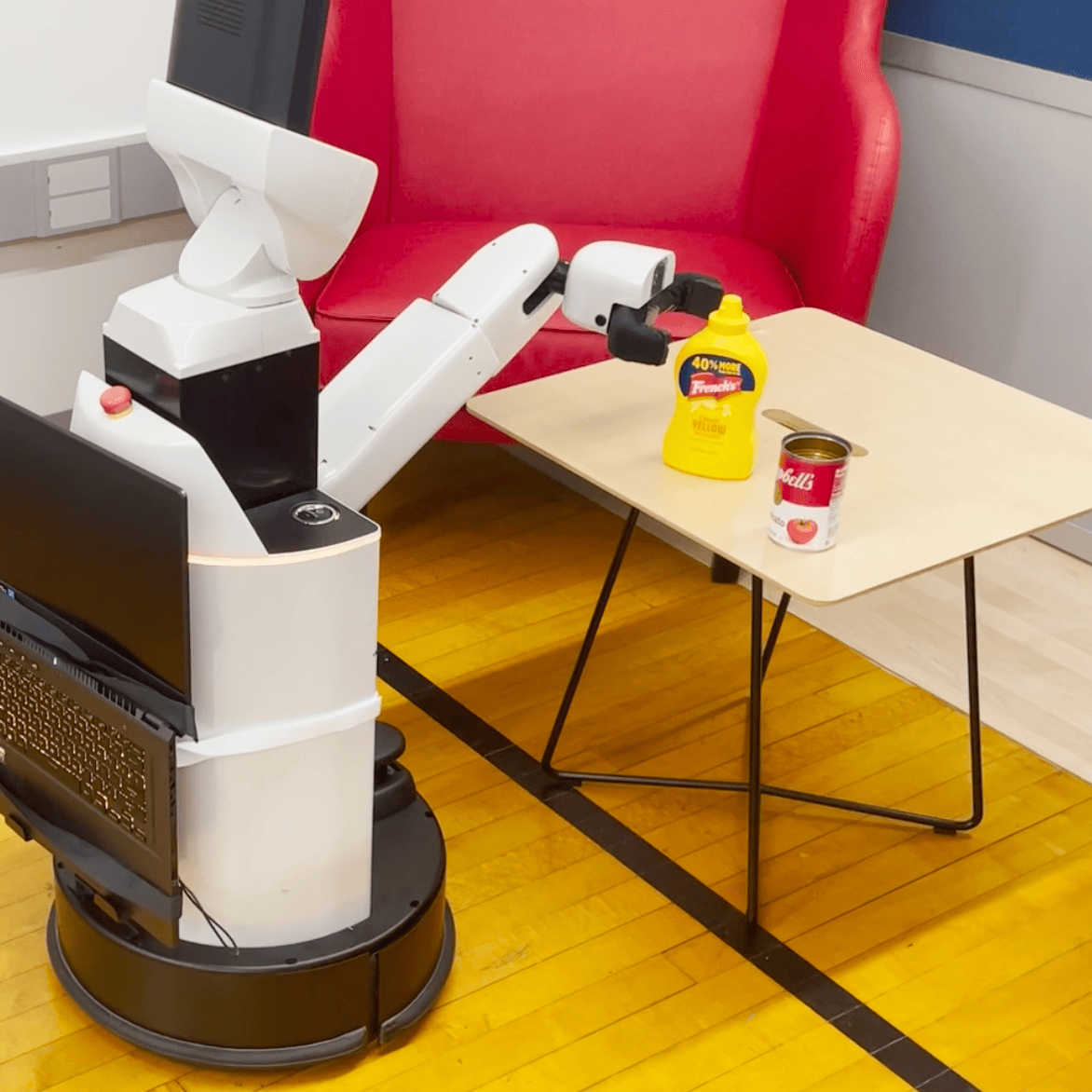}
         \caption{free gripper}
     \end{subfigure}
     \begin{subfigure}[b]{0.15\textwidth}
         \centering
         \includegraphics[width=\textwidth]{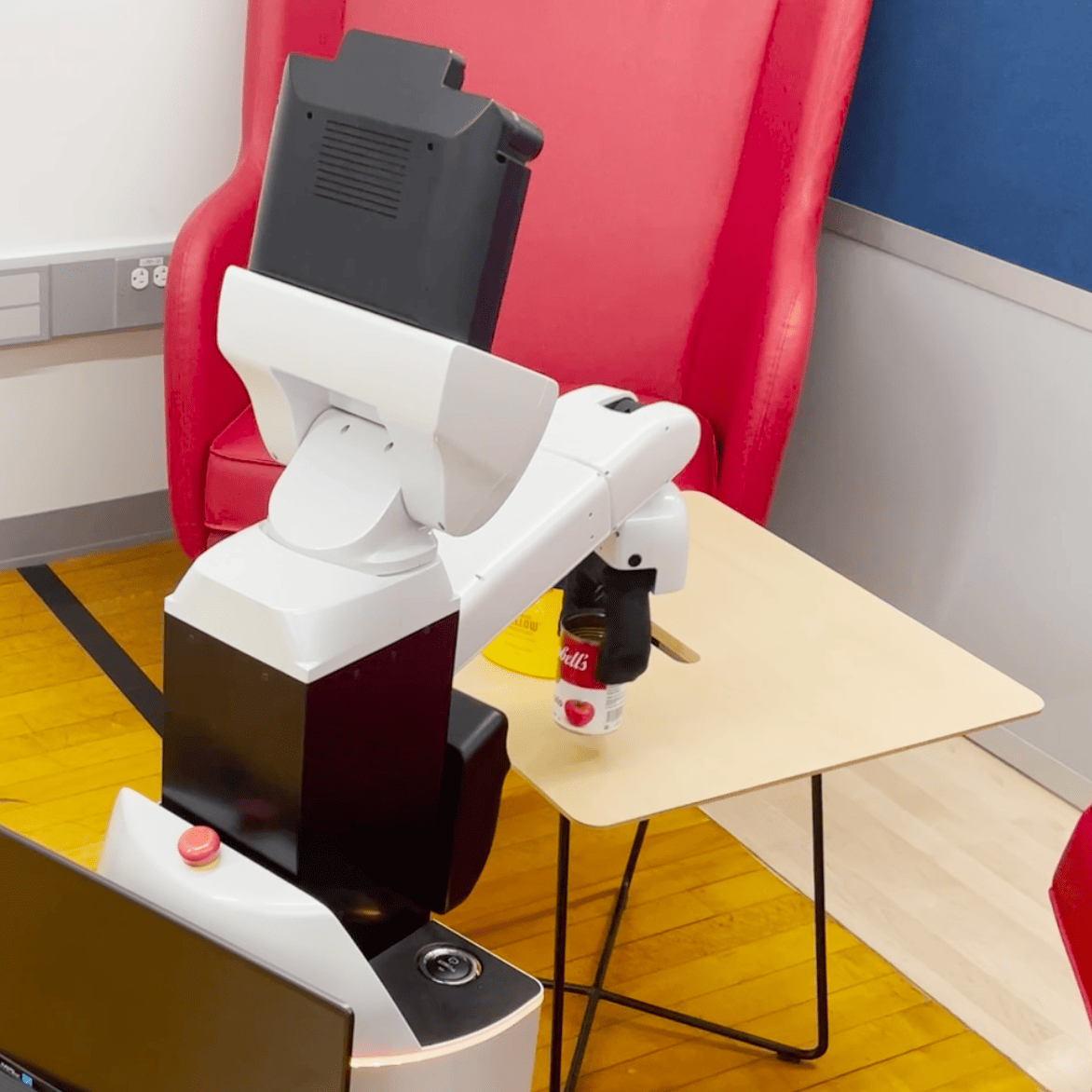}
         \caption{grasp soup can}
     \end{subfigure}
     \begin{subfigure}[b]{0.15\textwidth}
         \centering
         \includegraphics[width=\textwidth]{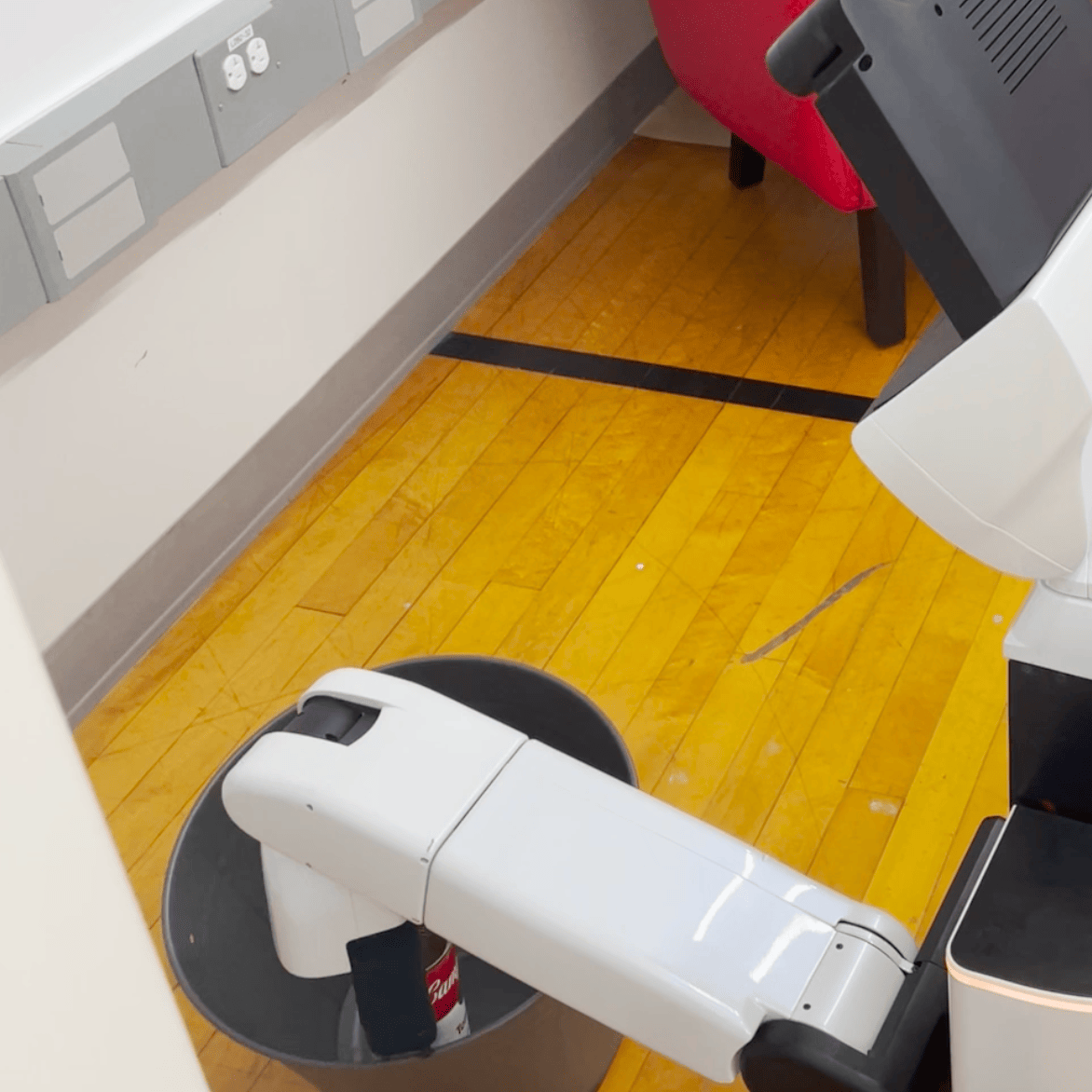}
         \caption{place soup can}
     \end{subfigure}
     \begin{subfigure}[b]{0.15\textwidth}
         \centering
         \includegraphics[width=\textwidth]{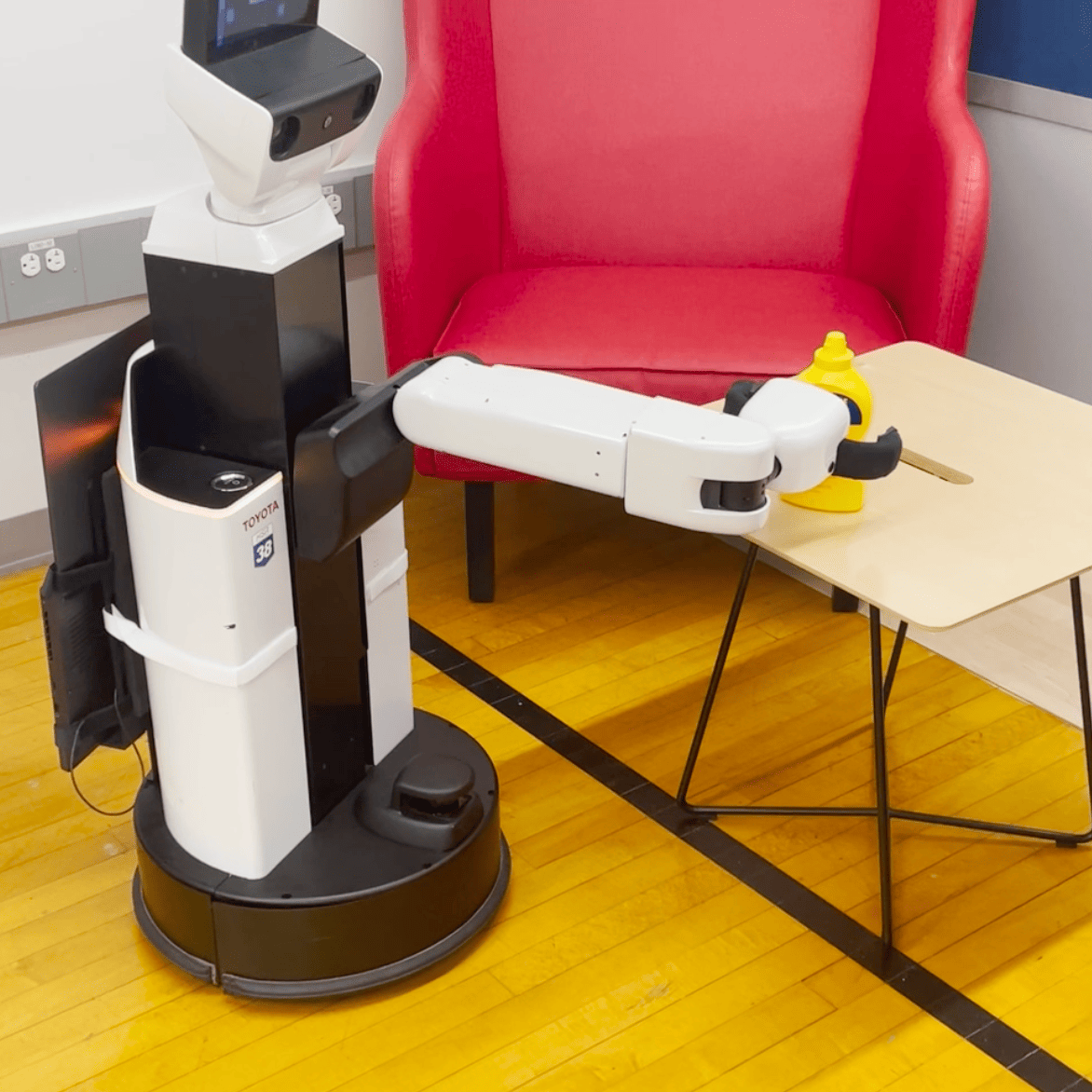}
         \caption{re-grasp bottle}
     \end{subfigure}
     \begin{subfigure}[b]{0.15\textwidth}
         \centering
         \includegraphics[width=\textwidth]{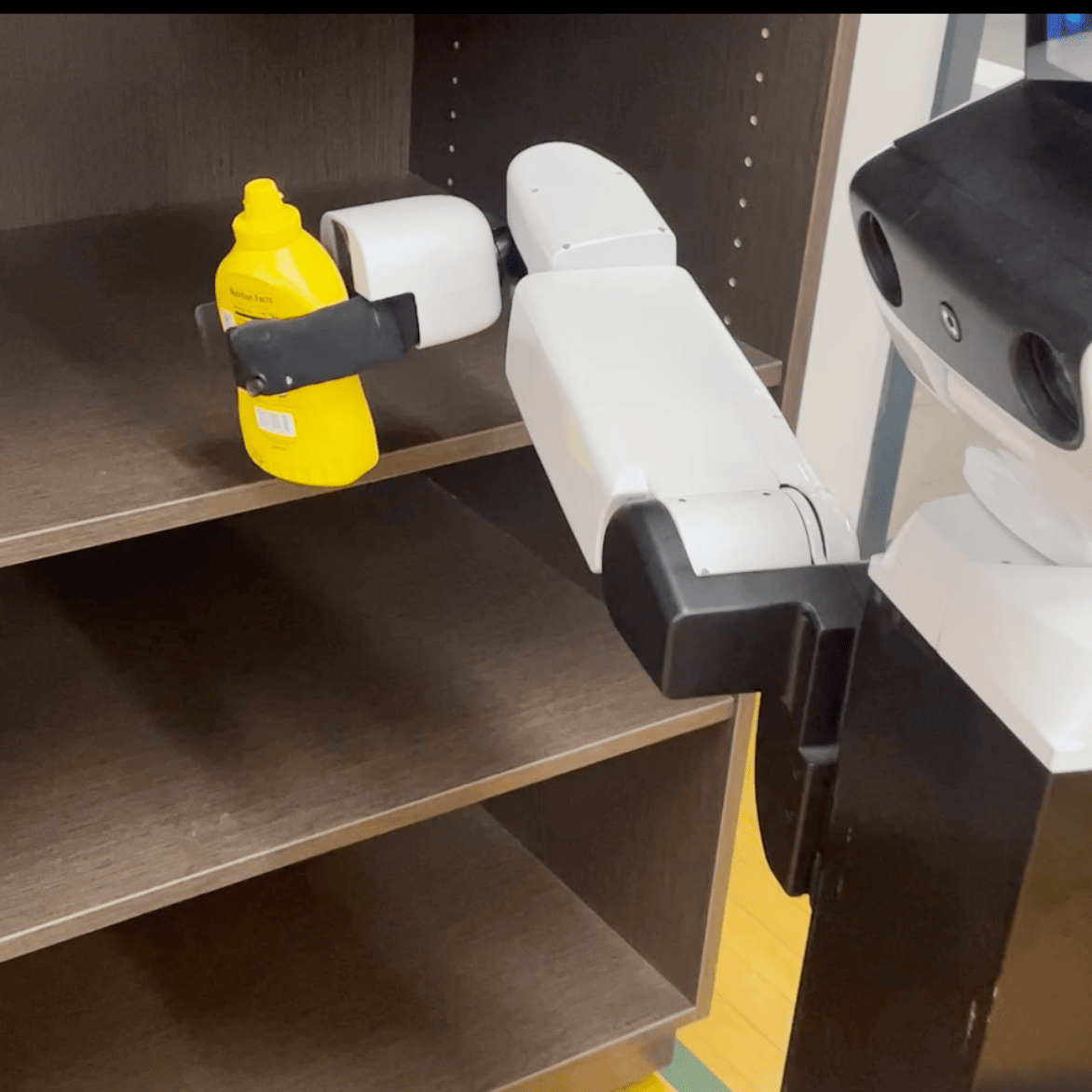}
         \caption{place bottle}
     \end{subfigure}
    \caption{Demonstration of the optimal tidy-up plan. The robot starts at the coffee table and 1) picks up the bottle, 2) navigates to a room with the side table and the recycle bin, 3) puts down the bottle, 4) grasps the soup can, 5) puts the soup can in the recycle bin, 6) re-grasps the bottle, 7) navigates to the kitchen, 8) places the bottle in the pantry. }
    \label{fig:demo}
    \vspace{-8pt}
\end{figure*}

\textbf{\textcolor{purple}{Findings (\lmsp{}):}}
\begin{enumerate}
    \item We observe that though \lmsp{} provides a plan in natural language for every problem, most of these plans are not feasible. The main reason is that \lmsp{} lacks the ability to reason about preconditions. 
    \item In most cases, \lmsp{} fails in the same way with or without the example plan as context. In particular, in the \blocksworld{} domain, \lmsp{} cannot keep track of properties like \textsc{on} and \textsc{clear}. In the \barman{} domain, \lmsp{}'s plans fail to clean shot glasses before using them again.  
    \item The hardest domains are the ones with complex spatial relationship. 
    The \lmsp{} methods (with or without context) completely fail at this type of problems. In the \floortile{} domain, \lmsp{} generates ``move right to tile\_0-4 and paint tile\_1-2 black" but the robot can only paint neighboring tiles. In \termes{} and \storage{}, \lmsp{} ignores the requirement that the robot cannot unload the block/crate at the same position it occupies.
    
    \item \lmtot{} calls the LLM at each tree node to provide a list of available actions, and then calls the LLM to evaluate each new path on the tree as a partial plan. We find that the LLM is able to give reasonable rankings on the partial plans, but it often fails to recognize whether the plan reaches the goal. \lmtot{} times out in most cases due to the large number of LLM calls, so it is not suitable for solving long-horizon problems.
\end{enumerate}

\textbf{\textcolor{purple}{Findings (\lmp{}):}}
\begin{enumerate}
    \item The proposed \lmp{} produces an optimal plan for the majority of problems. Most failed cases are due to mis-specified problem files, such as missing one of the initial conditions (e.g. leaving the tiles disconnected in \floortile{}), causing the planning problem to be unsolvable. 
    \item Without the context (i.e., an example problem and its corresponding problem PDDL), we observe that LLMs fail to produce correct problem PDDL files. 
    Therefore, the context is important for \lmp{} to work.
\end{enumerate}



\subsection{Robot Demonstration}
We verify that \lmp{} can efficiently solve realistic service robot problems by deploying it on a real robot tasked with tidying up a home. The user asks the robot to move a mustard bottle from the coffee table to the pantry, and throw away the empty soup can from the side table. Since the side table and the recycle bin are on the way from the coffee table to the pantry, the optimal plan is to take the mustard bottle to the side table, and re-grasp it after throwing away the soup can, with a total cost of 22. Fig. \ref{fig:demo} shows the optimal plan found by \lmp{}. Parts of the prompt and the generated PDDL are shown below. \lmsp{} outputs a sub-optimal plan which takes the bottle to the pantry first and travels back for the soup can, with a total cost of 31.  

\begin{tcolorbox}[
    standard jigsaw,
    title=Tidy-Up Problem PDDL Generated by \lmp{},
    opacityback=0,
]
\textbf{Problem (P):} You are a home robot with one gripper.
The distance between coffee table and side table is 10. The distance between coffee table and pantry is 20...
You are at the coffee table.
There is a mustard bottle...
Your goal is to move objects to their destinations...
 \\
\textbf{Problem PDDL generated by \lmp{}:}  \\
\PlanCode{(:objects
coffee-table side-table recycle-bin pantry - location
mustard-bottle soup-can - object)} \\
\PlanCode{(:init
(= (total-cost) 0)
(= (distance coffee-table side-table) 10)
(= (distance coffee-table pantry) 20)
...
(robot-at coffee-table)
(at mustard-bottle coffee-table)
(at soup-can side-table)
(hand-empty)
)} \\
\PlanCode{(:goal
(and
(at mustard-bottle pantry)
(at soup-can recycle-bin)))} 
\PlanCode{(:metric minimize (total-cost))
)}

\end{tcolorbox}

\section{Conclusion and Future Work}
In this work, we propose to leverage classical planners to empower large language models with optimal planning capabilities. The key design choice of the proposed \lmp{} framework is to focus LLMs on translating the planning problem from natural language to structured PDDL format. Moreover, we show that it is important to also make LLMs aware of a simple (problem, PDDL) pair as a demonstration (or the context) for in-context learning. Some interesting directions to further extend the \lmp{} framework include: 1) enabling the LLM to auto-detect when and how to apply \lmp{}; and 2) reducing  \lmp{}'s dependency on information by humans, potentially involving finetuning.

\clearpage


\bibliographystyle{IEEEtran}
\bibliography{references}

\begin{thebibliography}{10}
\providecommand{\url}[1]{#1}
\csname url@rmstyle\endcsname
\providecommand{\newblock}{\relax}
\providecommand{\bibinfo}[2]{#2}
\providecommand\BIBentrySTDinterwordspacing{\spaceskip=0pt\relax}
\providecommand\BIBentryALTinterwordstretchfactor{4}
\providecommand\BIBentryALTinterwordspacing{\spaceskip=\fontdimen2\font plus
\BIBentryALTinterwordstretchfactor\fontdimen3\font minus \fontdimen4\font\relax}
\providecommand\BIBforeignlanguage[2]{{%
\expandafter\ifx\csname l@#1\endcsname\relax
\typeout{** WARNING: IEEEtran.bst: No hyphenation pattern has been}%
\typeout{** loaded for the language `#1'. Using the pattern for}%
\typeout{** the default language instead.}%
\else
\language=\csname l@#1\endcsname
\fi
#2}}

\bibitem{weizenbaum1966eliza}
J.~Weizenbaum, ``Eliza—a computer program for the study of natural language communication between man and machine,'' \emph{Communications of the ACM}, vol.~9, no.~1, pp. 36--45, 1966.

\bibitem{openai2023gpt4}
OpenAI, ``Gpt-4 technical report,'' 2023.

\bibitem{vemprala2023chatgpt}
\BIBentryALTinterwordspacing
S.~Vemprala, R.~Bonatti, A.~Bucker, and A.~Kapoor, ``Chatgpt for robotics: Design principles and model abilities,'' Microsoft, Tech. Rep. MSR-TR-2023-8, February 2023. [Online]. Available: \url{https://www.microsoft.com/en-us/research/publication/chatgpt-for-robotics-design-principles-and-model-abilities/}
\BIBentrySTDinterwordspacing

\bibitem{mahowald2023dissociating}
K.~Mahowald, A.~A. Ivanova, I.~A. Blank, N.~Kanwisher, J.~B. Tenenbaum, and E.~Fedorenko, ``Dissociating language and thought in large language models: a cognitive perspective,'' \emph{arXiv preprint arXiv:2301.06627}, 2023.

\bibitem{lee2019mixout}
C.~Lee, K.~Cho, and W.~Kang, ``Mixout: Effective regularization to finetune large-scale pretrained language models,'' \emph{arXiv preprint arXiv:1909.11299}, 2019.

\bibitem{wei2021finetuned}
J.~Wei, M.~Bosma, V.~Y. Zhao, K.~Guu, A.~W. Yu, B.~Lester, N.~Du, A.~M. Dai, and Q.~V. Le, ``Finetuned language models are zero-shot learners,'' \emph{arXiv preprint arXiv:2109.01652}, 2021.

\bibitem{mcdermott1998pddl}
D.~McDermott, M.~Ghallab, A.~Howe, C.~Knoblock, A.~Ram, M.~Veloso, D.~Weld, and D.~Wilkins, ``Pddl-the planning domain definition language,'' 1998.

\bibitem{haslum2019introduction}
P.~Haslum, N.~Lipovetzky, D.~Magazzeni, and C.~Muise, ``An introduction to the planning domain definition language,'' \emph{Synthesis Lectures on Artificial Intelligence and Machine Learning}, vol.~13, no.~2, pp. 1--187, 2019.

\bibitem{valmeekam2022large}
K.~Valmeekam, A.~Olmo, S.~Sreedharan, and S.~Kambhampati, ``Large language models still can't plan (a benchmark for llms on planning and reasoning about change),'' \emph{arXiv preprint arXiv:2206.10498}, 2022.

\bibitem{brown2020language}
T.~Brown, B.~Mann, N.~Ryder, M.~Subbiah, J.~D. Kaplan, P.~Dhariwal, A.~Neelakantan, P.~Shyam, G.~Sastry, A.~Askell, \emph{et~al.}, ``Language models are few-shot learners,'' \emph{Advances in neural information processing systems}, vol.~33, pp. 1877--1901, 2020.

\bibitem{helmert2006fast}
M.~Helmert, ``The fast downward planning system,'' \emph{Journal of Artificial Intelligence Research}, vol.~26, pp. 191--246, 2006.

\bibitem{bylander1994ComputationalComplexityPropositional}
T.~Bylander, ``The computational complexity of propositional {{STRIPS}} planning,'' \emph{Artificial Intelligence}, vol.~69, no. 1-2, pp. 165--204, 1994.

\bibitem{mccarthy1963situations}
J.~McCarthy, ``Situations, actions, and causal laws,'' Stanford University Technical Report, Tech. Rep., 1963.

\bibitem{fikes1971strips}
R.~E. Fikes and N.~J. Nilsson, ``Strips: A new approach to the application of theorem proving to problem solving,'' \emph{Artificial intelligence}, vol.~2, no. 3-4, pp. 189--208, 1971.

\bibitem{nilsson1984shakey}
N.~J. Nilsson \emph{et~al.}, ``Shakey the robot,'' 1984.

\bibitem{carbonell1991prodigy}
J.~Carbonell, O.~Etzioni, Y.~Gil, R.~Joseph, C.~Knoblock, S.~Minton, and M.~Veloso, ``Prodigy: An integrated architecture for planning and learning,'' \emph{ACM SIGART Bulletin}, vol.~2, no.~4, pp. 51--55, 1991.

\bibitem{nau2003shop2}
D.~S. Nau, T.-C. Au, O.~Ilghami, U.~Kuter, J.~W. Murdock, D.~Wu, and F.~Yaman, ``Shop2: An htn planning system,'' \emph{Journal of artificial intelligence research}, 2003.

\bibitem{jiang2019task}
Y.-q. Jiang, S.-q. Zhang, P.~Khandelwal, and P.~Stone, ``Task planning in robotics: an empirical comparison of pddl-and asp-based systems,'' \emph{Frontiers of Information Technology \& Electronic Engineering}, vol.~20, pp. 363--373, 2019.

\bibitem{brewka2011answer}
G.~Brewka, T.~Eiter, and M.~Truszczy{\'n}ski, ``Answer set programming at a glance,'' \emph{Communications of the ACM}, vol.~54, no.~12, pp. 92--103, 2011.

\bibitem{lifschitz2002answer}
V.~Lifschitz, ``Answer set programming and plan generation,'' \emph{Artificial Intelligence}, vol. 138, no. 1-2, pp. 39--54, 2002.

\bibitem{fox2003pddl2}
M.~Fox and D.~Long, ``Pddl2. 1: An extension to pddl for expressing temporal planning domains,'' \emph{Journal of artificial intelligence research}, vol.~20, pp. 61--124, 2003.

\bibitem{zhang2015mobile}
S.~Zhang, F.~Yang, P.~Khandelwal, and P.~Stone, ``Mobile robot planning using action language bc with an abstraction hierarchy,'' in \emph{International Conference on Logic Programming and Nonmonotonic Reasoning}.\hskip 1em plus 0.5em minus 0.4em\relax Springer, 2015, pp. 502--516.

\bibitem{ding2020task}
Y.~Ding, X.~Zhang, X.~Zhan, and S.~Zhang, ``Task-motion planning for safe and efficient urban driving,'' in \emph{2020 IEEE/RSJ International Conference on Intelligent Robots and Systems (IROS)}, 2020.

\bibitem{jiang2019multi}
Y.~Jiang, H.~Yedidsion, S.~Zhang, G.~Sharon, and P.~Stone, ``Multi-robot planning with conflicts and synergies,'' \emph{Autonomous Robots}, vol.~43, no.~8, pp. 2011--2032, 2019.

\bibitem{lagriffoul2018platform}
F.~Lagriffoul, N.~T. Dantam, C.~Garrett, A.~Akbari, S.~Srivastava, and L.~E. Kavraki, ``Platform-independent benchmarks for task and motion planning,'' \emph{IEEE Robotics and Automation Letters}, vol.~3, no.~4, pp. 3765--3772, 2018.

\bibitem{kaelbling2013integrated}
L.~P. Kaelbling and T.~Lozano-P{\'e}rez, ``Integrated task and motion planning in belief space,'' \emph{The International Journal of Robotics Research}, vol.~32, no. 9-10, pp. 1194--1227, 2013.

\bibitem{devlin2018bert}
J.~Devlin, M.-W. Chang, K.~Lee, and K.~Toutanova, ``Bert: Pre-training of deep bidirectional transformers for language understanding,'' \emph{arXiv preprint arXiv:1810.04805}, 2018.

\bibitem{chen2021evaluating}
M.~Chen, J.~Tworek, H.~Jun, Q.~Yuan, H.~P. d.~O. Pinto, J.~Kaplan, H.~Edwards, Y.~Burda, N.~Joseph, G.~Brockman, \emph{et~al.}, ``Evaluating large language models trained on code,'' \emph{arXiv preprint arXiv:2107.03374}, 2021.

\bibitem{zhang2022opt}
S.~Zhang, S.~Roller, N.~Goyal, M.~Artetxe, M.~Chen, S.~Chen, C.~Dewan, M.~Diab, X.~Li, X.~V. Lin, \emph{et~al.}, ``Opt: Open pre-trained transformer language models,'' \emph{arXiv preprint arXiv:2205.01068}, 2022.

\bibitem{openai}
\BIBentryALTinterwordspacing
OpenAI, ``Chatgpt,'' Accessed: 2023-02-08, 2023, cit. on pp. 1, 16. [Online]. Available: \url{https://openai.com/blog/chatgpt/}
\BIBentrySTDinterwordspacing

\bibitem{touvron2023llama}
H.~Touvron, T.~Lavril, G.~Izacard, X.~Martinet, M.-A. Lachaux, T.~Lacroix, B.~Rozi{\`e}re, N.~Goyal, E.~Hambro, F.~Azhar, \emph{et~al.}, ``Llama: Open and efficient foundation language models,'' \emph{arXiv preprint arXiv:2302.13971}, 2023.

\bibitem{touvron2023llama2}
H.~Touvron, L.~Martin, K.~Stone, P.~Albert, A.~Almahairi, Y.~Babaei, N.~Bashlykov, S.~Batra, P.~Bhargava, S.~Bhosale, \emph{et~al.}, ``Llama 2: Open foundation and fine-tuned chat models,'' \emph{arXiv preprint arXiv:2307.09288}, 2023.

\bibitem{chowdhery2022palm}
A.~Chowdhery, S.~Narang, J.~Devlin, M.~Bosma, G.~Mishra, A.~Roberts, P.~Barham, H.~W. Chung, C.~Sutton, S.~Gehrmann, \emph{et~al.}, ``Palm: Scaling language modeling with pathways,'' \emph{arXiv preprint arXiv:2204.02311}, 2022.

\bibitem{ahn2022can}
M.~Ahn, A.~Brohan, N.~Brown, Y.~Chebotar, O.~Cortes, B.~David, C.~Finn, K.~Gopalakrishnan, K.~Hausman, A.~Herzog, \emph{et~al.}, ``Do as i can, not as i say: Grounding language in robotic affordances,'' \emph{arXiv preprint arXiv:2204.01691}, 2022.

\bibitem{ding2023task}
Y.~Ding, X.~Zhang, C.~Paxton, and S.~Zhang, ``Task and motion planning with large language models for object rearrangement,'' \emph{2023 IEEE/RSJ International Conference on Intelligent Robots and Systems (IROS)}, 2023.

\bibitem{driess2023palm}
D.~Driess, F.~Xia, M.~S. Sajjadi, C.~Lynch, A.~Chowdhery, B.~Ichter, A.~Wahid, J.~Tompson, Q.~Vuong, T.~Yu, \emph{et~al.}, ``Palm-e: An embodied multimodal language model,'' \emph{arXiv preprint arXiv:2303.03378}, 2023.

\bibitem{huang2022inner}
W.~Huang, F.~Xia, T.~Xiao, H.~Chan, J.~Liang, P.~Florence, A.~Zeng, J.~Tompson, I.~Mordatch, Y.~Chebotar, \emph{et~al.}, ``Inner monologue: Embodied reasoning through planning with language models,'' \emph{arXiv preprint arXiv:2207.05608}, 2022.

\bibitem{huang2022language}
W.~Huang, P.~Abbeel, D.~Pathak, and I.~Mordatch, ``Language models as zero-shot planners: Extracting actionable knowledge for embodied agents,'' in \emph{International Conference on Machine Learning}.\hskip 1em plus 0.5em minus 0.4em\relax PMLR, 2022, pp. 9118--9147.

\bibitem{kant2022housekeep}
Y.~Kant, A.~Ramachandran, S.~Yenamandra, I.~Gilitschenski, D.~Batra, A.~Szot, and H.~Agrawal, ``Housekeep: Tidying virtual households using commonsense reasoning,'' in \emph{Computer Vision--ECCV 2022: 17th European Conference, Tel Aviv, Israel, October 23--27, 2022, Proceedings, Part XXXIX}.\hskip 1em plus 0.5em minus 0.4em\relax Springer, 2022, pp. 355--373.

\bibitem{singh2022progprompt}
I.~Singh, V.~Blukis, A.~Mousavian, A.~Goyal, D.~Xu, J.~Tremblay, D.~Fox, J.~Thomason, and A.~Garg, ``Progprompt: Generating situated robot task plans using large language models,'' \emph{arXiv preprint arXiv:2209.11302}, 2022.

\bibitem{lin2023text2motion}
K.~Lin, C.~Agia, T.~Migimatsu, M.~Pavone, and J.~Bohg, ``Text2motion: From natural language instructions to feasible plans,'' \emph{arXiv preprint arXiv:2303.12153}, 2023.

\bibitem{yang2023automatonbased}
Y.~Yang, J.-R. Gaglione, C.~Neary, and U.~Topcu, ``Automaton-based representations of task knowledge from generative language models,'' \emph{arXiv preprint arXiv:2212.01944}, 2023.

\bibitem{ding2023integrating}
Y.~Ding, X.~Zhang, S.~Amiri, N.~Cao, H.~Yang, A.~Kaminski, C.~Esselink, and S.~Zhang, ``Integrating action knowledge and llms for task planning and situation handling in open worlds,'' \emph{arXiv preprint arXiv:2305.17590}, 2023.

\bibitem{ren2023robots}
A.~Z. Ren, A.~Dixit, A.~Bodrova, S.~Singh, S.~Tu, N.~Brown, P.~Xu, L.~Takayama, F.~Xia, J.~Varley, \emph{et~al.}, ``Robots that ask for help: Uncertainty alignment for large language model planners,'' \emph{arXiv preprint arXiv:2307.01928}, 2023.

\bibitem{chen2023autotamp}
Y.~Chen, J.~Arkin, Y.~Zhang, N.~Roy, and C.~Fan, ``Autotamp: Autoregressive task and motion planning with llms as translators and checkers,'' \emph{arXiv preprint arXiv:2306.06531}, 2023.

\bibitem{valmeekam2023planning}
K.~Valmeekam, S.~Sreedharan, M.~Marquez, A.~Olmo, and S.~Kambhampati, ``On the planning abilities of large language models (a critical investigation with a proposed benchmark),'' \emph{arXiv preprint arXiv:2302.06706}, 2023.

\bibitem{silver2022pddl}
\BIBentryALTinterwordspacing
T.~Silver, V.~Hariprasad, R.~S. Shuttleworth, N.~Kumar, T.~Lozano-P{\'e}rez, and L.~P. Kaelbling, ``{PDDL} planning with pretrained large language models,'' in \emph{NeurIPS 2022 Foundation Models for Decision Making Workshop}, 2022. [Online]. Available: \url{https://openreview.net/forum?id=1QMMUB4zfl}
\BIBentrySTDinterwordspacing

\bibitem{pallagani2022plansformer}
V.~Pallagani, B.~Muppasani, K.~Murugesan, F.~Rossi, L.~Horesh, B.~Srivastava, F.~Fabiano, and A.~Loreggia, ``Plansformer: Generating symbolic plans using transformers,'' \emph{arXiv preprint arXiv:2212.08681}, 2022.

\bibitem{arora2023learning}
D.~Arora and S.~Kambhampati, ``Learning and leveraging verifiers to improve planning capabilities of pre-trained language models,'' \emph{arXiv preprint arXiv:2305.17077}, 2023.

\bibitem{guan2023leveraging}
L.~Guan, K.~Valmeekam, S.~Sreedharan, and S.~Kambhampati, ``Leveraging pre-trained large language models to construct and utilize world models for model-based task planning,'' \emph{arXiv preprint arXiv:2305.14909}, 2023.

\bibitem{silver2023generalized}
T.~Silver, S.~Dan, K.~Srinivas, J.~B. Tenenbaum, L.~P. Kaelbling, and M.~Katz, ``Generalized planning in pddl domains with pretrained large language models,'' \emph{arXiv preprint arXiv:2305.11014}, 2023.

\bibitem{pallagani2023understanding}
V.~Pallagani, B.~Muppasani, K.~Murugesan, F.~Rossi, B.~Srivastava, L.~Horesh, F.~Fabiano, and A.~Loreggia, ``Understanding the capabilities of large language models for automated planning,'' \emph{arXiv preprint arXiv:2305.16151}, 2023.

\bibitem{valmeekam2023planningnew}
K.~Valmeekam, M.~Marquez, S.~Sreedharan, and S.~Kambhampati, ``On the planning abilities of large language models--a critical investigation,'' \emph{arXiv preprint arXiv:2305.15771}, 2023.

\bibitem{xie2023translating}
Y.~Xie, C.~Yu, T.~Zhu, J.~Bai, Z.~Gong, and H.~Soh, ``Translating natural language to planning goals with large-language models,'' \emph{arXiv preprint arXiv:2302.05128}, 2023.

\bibitem{hazra2023saycanpay}
R.~Hazra, P.~Z.~D. Martires, and L.~De~Raedt, ``Saycanpay: Heuristic planning with large language models using learnable domain knowledge,'' \emph{arXiv preprint arXiv:2308.12682}, 2023.

\bibitem{rana2023sayplan}
K.~Rana, J.~Haviland, S.~Garg, J.~Abou-Chakra, I.~Reid, and N.~Suenderhauf, ``Sayplan: Grounding large language models using 3d scene graphs for scalable task planning,'' \emph{arXiv preprint arXiv:2307.06135}, 2023.

\bibitem{zhou2023isr}
Z.~Zhou, J.~Song, K.~Yao, Z.~Shu, and L.~Ma, ``Isr-llm: Iterative self-refined large language model for long-horizon sequential task planning,'' \emph{arXiv preprint arXiv:2308.13724}, 2023.

\bibitem{wang2023describe}
Z.~Wang, S.~Cai, A.~Liu, X.~Ma, and Y.~Liang, ``Describe, explain, plan and select: Interactive planning with large language models enables open-world multi-task agents,'' \emph{arXiv preprint arXiv:2302.01560}, 2023.

\bibitem{nakano2021webgpt}
R.~Nakano, J.~Hilton, S.~Balaji, J.~Wu, L.~Ouyang, C.~Kim, C.~Hesse, S.~Jain, V.~Kosaraju, W.~Saunders, \emph{et~al.}, ``Webgpt: Browser-assisted question-answering with human feedback,'' \emph{arXiv preprint arXiv:2112.09332}, 2021.

\bibitem{lazaridou2022internet}
A.~Lazaridou, E.~Gribovskaya, W.~Stokowiec, and N.~Grigorev, ``Internet-augmented language models through few-shot prompting for open-domain question answering,'' \emph{arXiv preprint arXiv:2203.05115}, 2022.

\bibitem{madaan2023memoryassisted}
A.~Madaan, N.~Tandon, P.~Clark, and Y.~Yang, ``Memory-assisted prompt editing to improve gpt-3 after deployment,'' 2023.

\bibitem{shi2023replug}
W.~Shi, S.~Min, M.~Yasunaga, M.~Seo, R.~James, M.~Lewis, L.~Zettlemoyer, and W.-t. Yih, ``Replug: Retrieval-augmented black-box language models,'' \emph{arXiv preprint arXiv:2301.12652}, 2023.

\bibitem{chen2022program}
W.~Chen, X.~Ma, X.~Wang, and W.~W. Cohen, ``Program of thoughts prompting: Disentangling computation from reasoning for numerical reasoning tasks,'' \emph{arXiv preprint arXiv:2211.12588}, 2022.

\bibitem{gao2022pal}
L.~Gao, A.~Madaan, S.~Zhou, U.~Alon, P.~Liu, Y.~Yang, J.~Callan, and G.~Neubig, ``Pal: Program-aided language models,'' \emph{arXiv preprint arXiv:2211.10435}, 2022.

\bibitem{schick2023toolformer}
T.~Schick, J.~Dwivedi-Yu, R.~Dess{\`\i}, R.~Raileanu, M.~Lomeli, L.~Zettlemoyer, N.~Cancedda, and T.~Scialom, ``Toolformer: Language models can teach themselves to use tools,'' \emph{arXiv preprint arXiv:2302.04761}, 2023.

\bibitem{lyu2023faithful}
Q.~Lyu, S.~Havaldar, A.~Stein, L.~Zhang, D.~Rao, E.~Wong, M.~Apidianaki, and C.~Callison-Burch, ``Faithful chain-of-thought reasoning,'' \emph{arXiv preprint arXiv:2301.13379}, 2023.

\bibitem{seipp-et-al-zenodo2022}
J.~Seipp, {\'A}.~Torralba, and J.~Hoffmann, ``{PDDL} generators,'' \url{https://doi.org/10.5281/zenodo.6382173}, 2022.

\bibitem{yao2023tree}
S.~Yao, D.~Yu, J.~Zhao, I.~Shafran, T.~L. Griffiths, Y.~Cao, and K.~Narasimhan, ``Tree of thoughts: Deliberate problem solving with large language models,'' \emph{arXiv preprint arXiv:2305.10601}, 2023.

\end{thebibliography}


\end{document}